 \documentclass[pmlr,twocolumn,10pt]{jmlr} 

\usepackage{booktabs}
\usepackage{siunitx}
\usepackage{placeins}

\usepackage[switch]{lineno}

\jmlrvolume{259}
\jmlryear{2024}
\jmlrsubmitted{LEAVE UNSET}
\jmlrpublished{LEAVE UNSET}
\jmlrworkshop{Machine Learning for Health (ML4H) 2024} 

\title[Enhancing Trust in Clinically Significant Prostate Cancer Prediction with Multiple MRI Modalities]{Enhancing Trust in Clinically Significant Prostate Cancer Prediction with Multiple Magnetic Resonance Imaging Modalities}

\author{%
  \Name{Benjamin Ng} \Email{benjamin.ng@uwaterloo.ca}\\
  \addr University of Waterloo, Canada
  \AND
  \Name{Chi-en Amy Tai}
  \Email{amy.tai@uwaterloo.ca}\\
  \addr University of Waterloo, Canada
  \AND
  \Name{E. Zhixuan Zeng}
  \Email{emilyzhixuan.zeng@uwaterloo.ca}\\
  \addr University of Waterloo, Canada
  \AND
  \Name{Alexander Wong} \Email{alexander.wong@uwaterloo.ca}\\
  \addr University of Waterloo, Canada
  \vspace{-1cm}
}

\begin{document}

\maketitle

\begin{abstract}
In the United States, prostate cancer is the second leading cause of deaths in males with a predicted 35,250 deaths in 2024. However, most diagnoses are non-lethal and deemed clinically insignificant which means that the patient will likely not be impacted by the cancer over their lifetime. As a result, numerous research studies have explored the accuracy of predicting clinical significance of prostate cancer based on magnetic resonance imaging (MRI) modalities and deep neural networks. Despite their high performance, these models are not trusted by most clinical scientists as they are trained solely on a single modality whereas clinical scientists often use multiple magnetic resonance imaging modalities during their diagnosis. In this paper, we investigate combining multiple MRI modalities to train a deep learning model to enhance trust in the models for clinically significant prostate cancer prediction. The promising performance and proposed training pipeline showcase the benefits of incorporating multiple MRI modalities for enhanced trust and accuracy.  

\end{abstract}

\paragraph*{Data and Code Availability}
We use the SPIE-AAPM-NCI PROSTATEx Challenges, PROSTATEx\_masks, and The Cancer Imaging Archive (TCIA) datasets~\citep{Litjens2014,Litjens2017,Cuocolo2021,Clark2013}. The data is available online and is available to other researchers. The code is available at \url{https://github.com/catai9/multiple-modality-prostate-cancer-prediction}.

\paragraph*{Institutional Review Board (IRB)}
Our research does not require IRB approval. 

\section{Introduction}
\label{sec:intro}
In the United States, prostate cancer is the second leading cause of deaths in males with a predicted 35,250 deaths in 2024~\citep{seer_prostate_cancer}. However, most diagnoses are non-lethal and deemed clinically insignificant which means that the patient will likely not be impacted by the cancer over their lifetime~\citep{Shaw2014-ci}. Thus, clinicians must distinguish between clinically significant and insignificant cancers to provide essential treatment when necessary while also avoiding overdiagnosis and overtreatment.

In the past decade, advances in machine learning, image classification, and semantic segmentation have made it possible to integrate deep learning into the clinical workflow for cancer diagnosis. 
Numerous research studies have explored the accuracy of predicting the clinical significance of prostate cancer based on magnetic resonance imaging (MRI) modalities and deep neural networks~\citep{li2022machine}. Examples include ~\citet{yoo2019prostate} which utilized diffusion-weighted magnetic resonance imaging (DWI) for cancer classification, and ~\citet{tai2024enhancing}, which focuses on the T2-weighted (T2w) modality. ~\citet{wang2018automated} utilized both T2w and Apparent Diffusion Coefficient (ADC) images to localize the prostate and detect clinically significant prostate cancer, but did not use DWI.

Despite their high performance, these models are not trusted by most clinical scientists. One reason is the lack of transparency and explanations provided by the deep learning models~\citep{bluemke2020assessing, european2019radiologist, jia2020clinical}. Additionally, clinical scientists often use multiple magnetic resonance imaging modalities for their diagnosis~\citep{steiger2016prostate}, whereas most prediction models only consider a single or limited set of modalities.

\begin{figure*}[htbp]
    \centering
    \includegraphics[width=\linewidth]{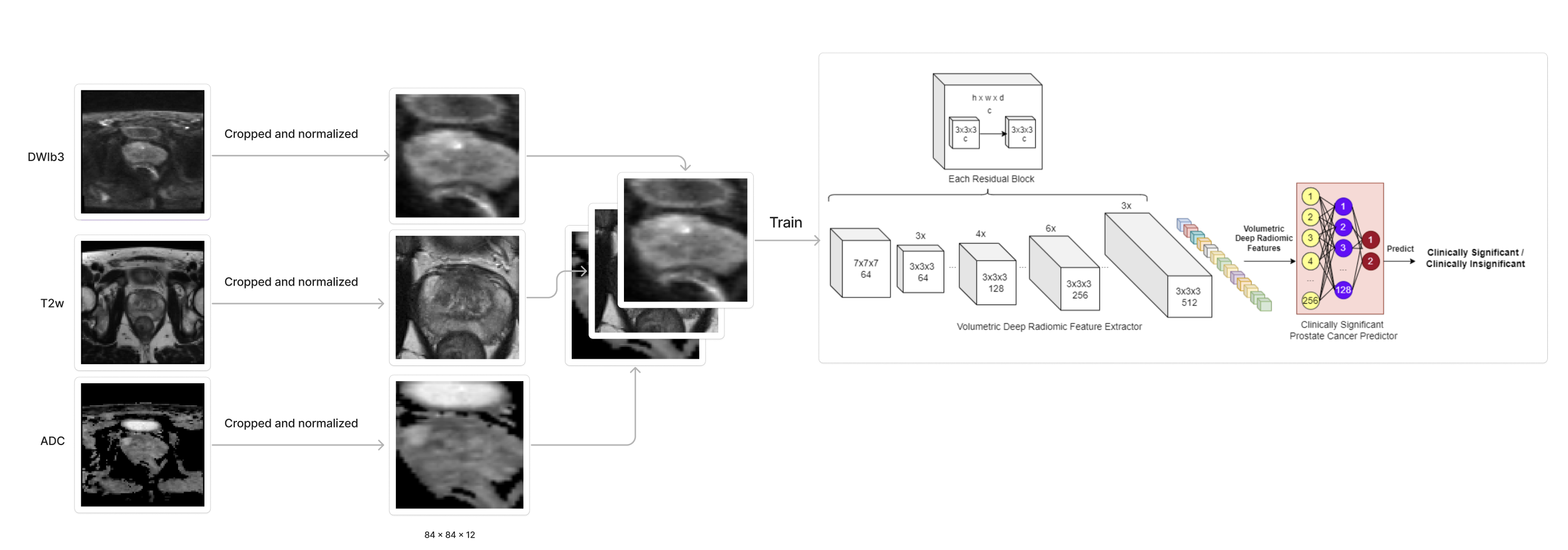}
    \caption{Preprocessing and training flow chart for 3-modality prostate cancer clinical significance model with the volumetric deep radiomic features and model portion adapted from~\citep{tai2024enhancing}.}
    \label{fig:flowchart}
\end{figure*}

In this paper, we investigate combining multiple MRI modalities to train a deep learning model to enhance trust in the models for clinically significant prostate cancer prediction. We further analyze the results using Explainable AI (XAI) techniques, which are analyzed and verified by a clinical scientist.

\section{Methodology}
\label{sec:method}
In this study, we first use XAI to gain insight into a high-performing model that was recently published~\citep{tai2024enhancing}, which we will refer to as the Tai and Wong Model. Building on these insights, we then guide model improvements and conduct comparative analysis of single versus multi-modal MRI approaches for machine learning.

\subsection{Explainable AI}
To validate a model’s performance, Grad-Cam++ implemented with M3D MedCam~\citep{2007.00453} was utilized to generate 3D classification and segmentation attention maps. The attention maps were extracted from the last applicable layer~\citep{chattopadhay2018grad}.

To generate the summed attention maps, leave-one-out cross-validation was first used to organize each patient prostate MRI volume into one of the four output categories: true positive, true negative, false positive or false negative. Then, for every output category, a summed attention map was generated by passing each volume in that category through the model and summing the resulting attention maps. Finally, the outputs were visualized using the jet colormap, as seen in Figure \ref{fig:overallResults}.

\subsection{Model Training}
As transfer learning was shown to perform best in this task~\citep{tai2023enhancing}, the model employed for training was the MONAI ResNet-34~\citep{cardoso2022monai} with initial weights from breast cancer grade prediction~\citep{tai2023enhancing}. In this study, we also use the cohort of 200 patients from the SPIE-AAPM-NCI PROSTATEx Challenges, PROSTATEx\_masks, and The Cancer Imaging Archive (TCIA) datasets for comparison consistency~\citep{Litjens2014,Litjens2017,Cuocolo2021,Clark2013}. Model training was conducted with a learning rate of 0.001 over 40 epochs, with no data augmentation or transformations applied. 

Leave-one-out cross-validation~\citep{cheng2017efficient} was employed to split the dataset, ensuring that each sample served once as the validation set while the remaining data were used for training, thus maximizing the use of the limited dataset. The cross-entropy loss function~\citep{goodfellow2016deep} was utilized to compute the error between the predicted and actual labels. The model output is binary, with a prediction of 1 indicating a clinically significant case of prostate cancer and 0 indicating the clinically insignificant prostate cancer. Model predictions were compared with expert medical evaluations to assess the model’s accuracy.

\subsection{Multiple MRI Modality Training}
The three MRI modalities studied were DWI with a b-value of 800 (referred to as DWIb3), T2w, and ADC. For each modality, the 3D volumes were preprocessed by standardizing to 12 slices and cropping each slice to focus on the prostate region. A composite 3D volume was then constructed by stacking the corresponding slices from the three modalities. The process of preprocessing and training this 3-modality model is shown in Figure~\ref{fig:flowchart}.

\section{Results}
\label{sec:results}

\subsection{XAI on the Tai and Wong Model}
\label{sec:problems}
\begin{figure}[ht]
    \centering
    \includegraphics[width=0.9\linewidth]{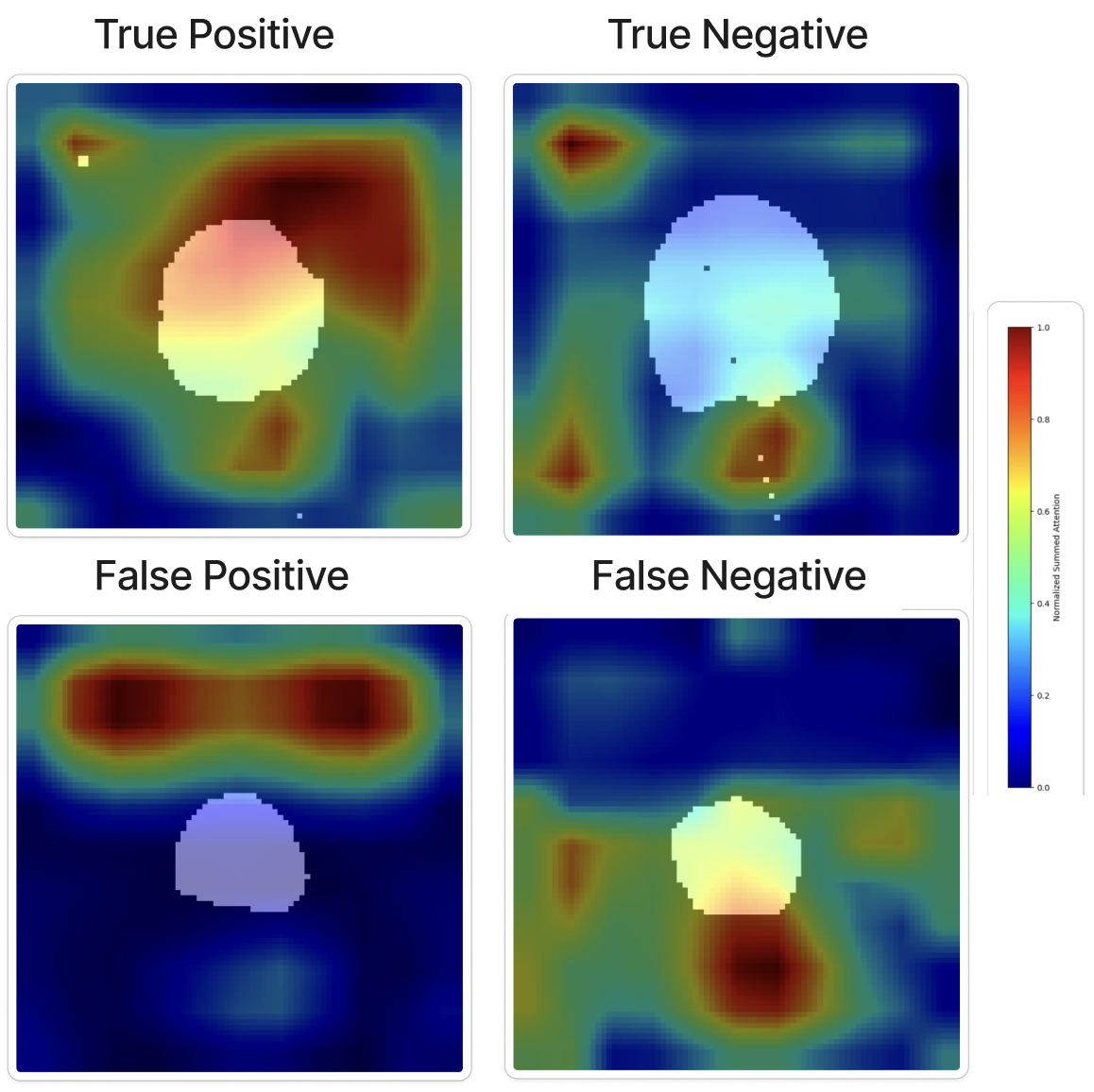}
    \caption{Summed Grad-CAM++ attention map and prostate mask for the Tai and Wong model~\citep{tai2024enhancing} (used to identify unintended behaviour of T2w-only model and potential overfitting).}
    \label{fig:overallResults}
\end{figure}

This model was trained using only the T2-weighted (T2w) modality and achieved a leave-one-out cross-validation accuracy of 97.5\%~\citep{tai2024enhancing}. However, following consultation with a medical professional and further analysis using M3D MedCam, this model was determined to be suboptimal, likely due to overfitting on the volumetric data. This was identified after creating a summed attention map for Grad-Cam++ for each of the patients in the output categories (true positive, true negative, false positive and false negative). The expected result is that the highlighting would be primarily around the prostate mask region, i.e., indicating that the model is looking at the prostate to determine clinical significance. However, as demonstrated in Figure~\ref{fig:overallResults}, the highlighting was sporadic and not centered around the prostate region. 

\subsection{Guided Model Improvement}
\begin{figure}
    \hfill
    \subfigure[Example Slice]{\includegraphics[width=0.43\linewidth]{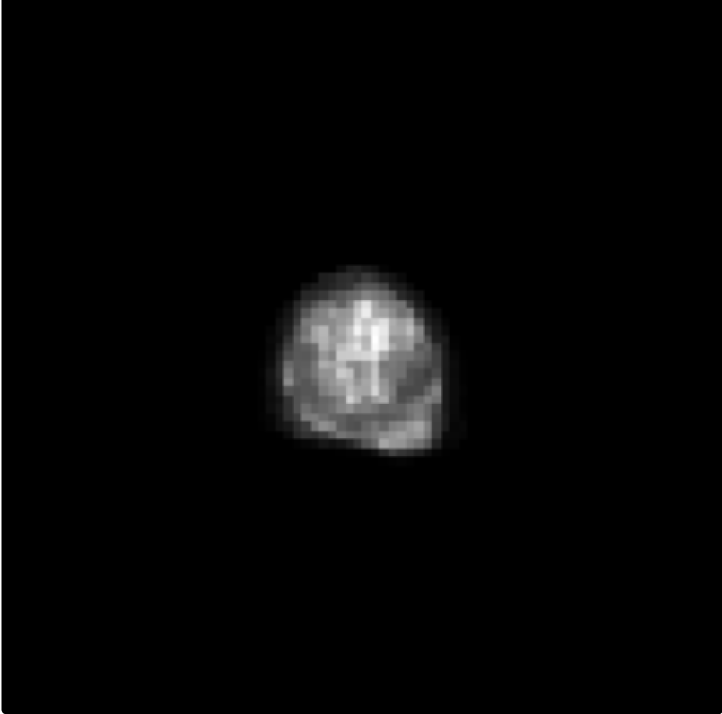}}
    \hfill
    \subfigure[Grad-Cam++ overlay]{\includegraphics[width=0.43\linewidth]{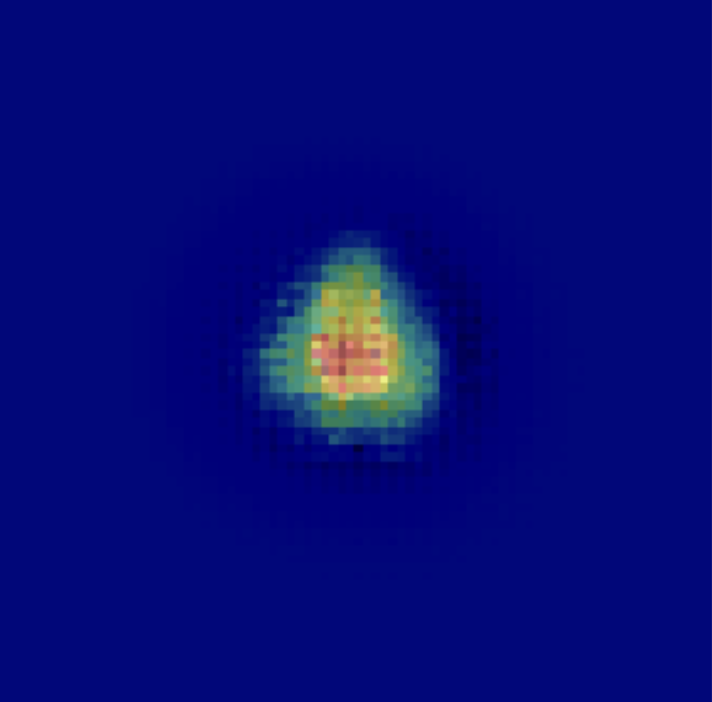}}
    \hfill
    \caption{Example slice for the single modality (T2w) model trained on images with masked prostate region with the associated Grad-Cam++ result.}
    \label{fig:singleModalityXAIResults}
    \vspace{-5pt}
\end{figure}

The XAI insights into the Tai and Wong model guided model improvement through explicit focusing around the prostate. Specifically we preprocessed the T2w modality patient volumes and masked areas outside of the prostate region. As a result, only the prostate region was visible and the rest of the image is black as demonstrated in the left side of Figure~\ref{fig:singleModalityXAIResults}. We then trained a model using these processed patient volumes. Although improving the outcome of the XAI (Figure~\ref{fig:singleModalityXAIResults} right), this adjustment resulted in a reduced accuracy of 65\%. This lower performance is attributed to the poor resolution of the images which led to insufficient information provided for model training. 

\subsection{3-Modality Model Results}
\begin{table}
\caption{Accuracy (Acc.), Sensitivity (Sen.), Specificity (Spec.), and F1 score comparison for (1) the Tai and Wong model (trained only on T2w)~\citep{tai2023enhancing}, (2) the single modality model (masked to the prostate region), and (3) the 3-modality model with cropping to the prostate region.}
    \centering
    \begin{tabular}{l c c c c}
        \hline
        Model & Acc. & Sen. & Spec. & F1 score \\ \hline
        (1) & 97.5\% & 94.3\% & 99.2\% & 96.3\% \\
        (2) & 65.0\% & 0.0\% & 100.0\% & 0.0\%\\
        (3) & 85.0\% & 84.3\% & 85.4\% & 79.7\%\\
    \hline
    \end{tabular}
    \label{tab:model-accuracy}
\end{table}

\begin{figure*}[htbp]
    \centering
    \includegraphics[width=\linewidth]{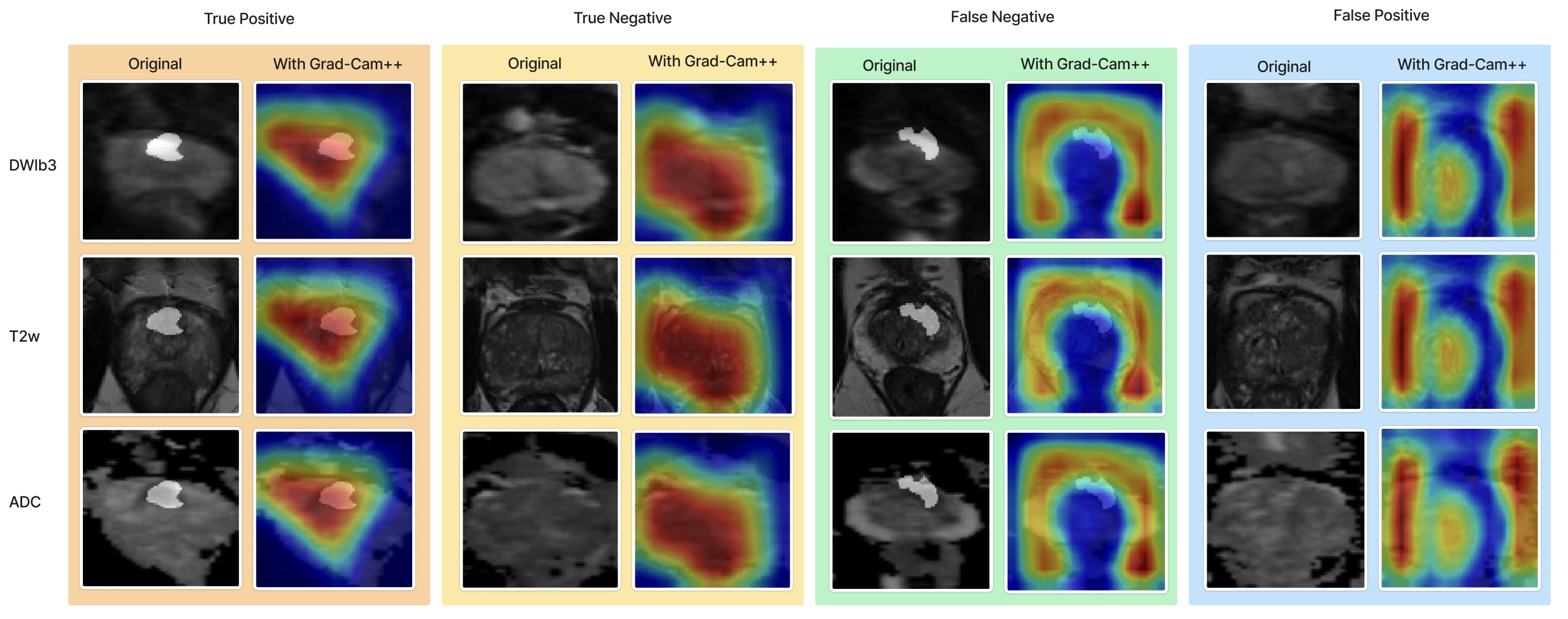}
    \caption{Example patient original (source image) slice with lesion mask (white highlighted region) vs Grad-Cam++ attention map for the 3-modality trained model.}
    \label{fig:outputCategoryResults}
    \vspace{-10pt}
\end{figure*}
 
The 3-modality model involved combining the three modalities and cropping the slices around the prostate. As shown in Table~\ref{tab:model-accuracy}, this model performed superior to the single modality masked. The 3-modality model demonstrated a leave-one-out cross-validation accuracy of 85\% in accurately identifying cases of clinically significant prostate cancer. Though the performance was not as impressive as the Tai and Wong model, the 3-modality model had better and more interpretable XAI results. Figure~\ref{fig:outputCategoryResults} illustrates the model’s prediction results, classified into four distinct categories.

For the true positive outcome, the highlighted regions coincided with the lesion mask (which represents the area where the clinically significant prostate should be identified), while the false negative outcome did not. This result aligns with expected behaviour as the model is expected to focus on the lesion mask within the prostate region to make accurate predictions. Conversely, in true negative cases, the model correctly identified the absence of lesions by examining the central region of the image where the lesion mask would be present. This observation also supports the conclusion that the model is focusing on the right regions.

\section{Conclusion}
\label{sec:concl}

This paper leverages XAI to provide two insights about prostate model training enhancements: cropping to focus around the prostate region and using multiple modalities. Initially, XAI showed issues in the Tai and Wong model because the attention map did not focus on the prostate. However, by addressing these issues through the aforementioned methods, we achieved more interpretable XAI results. This outcome demonstrates that combining multiple MRI modalities in deep learning model training has the potential to foster greater trust in clinically significant prostate cancer prediction. Furthermore, the strong accuracy of the multi-modality training pipeline highlights the effectiveness of this approach. Future work includes expanding the study with other prostate cancer datasets and studying the XAI results for other high-performing models in the cancer domain to analyze their explainability. 


\bibliography{references}

\end{document}